\documentclass[runningheads]{llncs}
\usepackage{graphicx}
\usepackage{dirtytalk}
\usepackage{paralist}
\usepackage{tcolorbox}
\usepackage{float}
\usepackage{cite}
\usepackage{orcidlink}
\let\llncssubparagraph\subparagraph
\let\subparagraph\paragraph
\usepackage[compact]{titlesec} 
\let\subparagraph\llncssubparagraph
\begin{document}
\title{
Towards Cognitive Bots: \\
Architectural Research Challenges
}
\author{
Habtom Kahsay Gidey\inst{1}\orcidlink{0000-0001-5802-2606} \and
Peter Hillmann\inst{1}\orcidlink{0000-0003-4346-4510} \and
Andreas Karcher\inst{1} \and
Alois Knoll\inst{2}\orcidlink{0000-0003-4840-076X}
}
\authorrunning{H.K Gidey et al.}
%
\institute{
Universität der Bundeswehr München, Germany \\
\email{\{habtom.gidey, peter.hillmann, andreas.karcher\}@unibw.de} \and
Technische Universität München, München, Germany\\
\email{\{knoll\}@in.tum.de}
}
\maketitle              
\begin{sloppypar}
\begin{abstract}
Software bots operating in multiple virtual digital platforms must understand the platforms' affordances and behave like human users. 
Platform affordances or features differ from one application platform to another or through a life cycle, requiring such bots to be adaptable. 
Moreover, bots in such platforms could cooperate with humans or other software agents for work or to learn specific behavior patterns.
However, present-day bots, particularly chatbots, other than language processing and prediction, are far from reaching a human user's behavior level within complex business information systems. 
They lack the cognitive capabilities to sense and act in such virtual environments, rendering their development a challenge to artificial general intelligence research. 
In this study, we problematize and investigate assumptions in conceptualizing software bot architecture by directing attention to significant architectural research challenges in developing cognitive bots endowed with complex behavior for operation on information systems.  
As an outlook, we propose alternate architectural assumptions to consider in future bot design and bot development frameworks.
\end{abstract}
%
%
%
\section{Introduction}\label{introduction}
Bots are software agents that operate in digital virtual environments~\cite{lebeuf2018taxonomy,etzioni1993intelligence}. 
An example scenario would be a~\say{\textit{user-like}} bot that could access web platforms and behave like a human user. 
Ideally, such a bot could autonomously sense and understand the platforms' affordances. 
Affordances in digital spaces are, for example, interaction possibilities and functionalities on the web, in software services, or on web application platforms~\cite{charpenay6,lemee2022signifiers}. 
The bot would recognize and understand the differences and variability between different environments' affordances. 
If a platform or service has extensions to physical bodies or devices, as in the Web of Things (WoT), it would also have control of or possibilities to interact with an outer web or service application world.

Ideally, a bot could also be independent of a specific platform. 
A user-like social bot, for instance, would be able to recognize and understand social networks and act to influence or engage in belief sharing on any social platform. 
It would also adjust with the changes and uncertainty of the affordances in a social media environment, such as when hypermedia interactivity features and functionalities change. 
Such a bot could also learn and develop to derive its goals and intentions from these digital microenvironments and take goal-directed targeted action to achieve them~\cite{goertzel2017toward}. 
Such bots could also communicate and cooperate with other user agents, humans, or bots to collaborate and socialize for collective understanding and behavior. 

The example scenarios described above convey desiderata of perception and action in bots, similar to how a human user would perceive and act in digital spaces. 
To date, bots are incapable of the essential cognitive skills required to engage in such activity since this would entail complex visual recognition, language understanding, and the employment of advanced cognitive models. 
Instead, most bots are either conversational interfaces or question-and-answer knowledge agents~\cite{lebeuf2018taxonomy}. 
Others only perform automated repetitive tasks based on pre-given rules, lacking autonomy and other advanced cognitive skills~\cite{ivanvcic2019robotic,engel2022cognitive}. 
The problems of realizing these desiderata are, therefore, complex and challenging~\cite{russell2010artificial,yampolskiy2012ai}. 
Solutions must address different areas, such as transduction and autonomous action, to achieve advanced generalizable intelligent behavior~\cite{goertzel2014engineeringPart1,vernon2016desiderata}.

Problems spanning diverse domains require architectural solutions. 
Accordingly, these challenges also necessitate that researchers address the structural and dynamic elements of such systems from an architectural perspective.~\cite{goertzel2014engineeringPart2,rosenbloom2023thoughts,lieto2018role}. 
For this reason, this paper outlines the architectural research agendas to address the challenges in conceptualizing and developing a cognitive bot with generalizable intelligence. 

The paper is divided into sections discussing each of the research challenges. 
In Sect.~\ref{transduction}, we discuss the challenges related to efforts and possible directions in enabling bots to sense and understand web platforms. 
Next, Sect.~\ref{behavior} describes the challenges related to developing advanced cognitive models in software bots. 
Sect.~\ref{communication} and~\ref{cooperation} discuss the research issues in bot communication and cooperation, respectively. 
The remaining two sections provide general discussions on bot ethics and trust and conclude the research agenda.
%
%
%
\section{The Transduction Problem}\label{transduction}
Web platforms can be seen as distinct microenvironments within digital microcosms~\cite{fountain1990microcosm}. 
They offer a microhabitat for their users' diverse digital experiences. 
These experiences mainly transpire from the elements of interaction and action, or the hypermedia, within web environments~\cite{fountain1990microcosm,nelson1965complex}. 
Hypermedia connects and extends the user experience, linking to further dimensions of the web-worlds, which means more pages and interactive elements from the user's perspective.
The interaction elements are considered affordances in the digital space~\cite{charpenay6,lemee2022signifiers}, analogous to the biological concept of affordances from environmental psychology~\cite{gibson1977theory}. 
Affordances can also be accompanied by signifiers. 
Signifiers reveal or indicate possibilities for actions associated with affordances~\cite{lemee2022signifiers,vachtsevanou2023signifiers}. 
An example on the web would be a button affording a click action and a text signifier hinting~\say{Click to submit}. 
A human user understands this web environment, its content, and its affordances, and navigates reasonably easily. 
However, enabling software bots to understand this digital environment and its affordances the way human users do is a challenging task. 
It is a complex problem of translating and mapping perception to action, i.e., the transduction problem~\cite{wooldridge2009introduction,brooks1991intelligence}.

Today, there are different approaches to this problem. 
The first category of approaches provides knowledge about the environment for different levels of observability using APIs or knowledge descriptions. 
With API-based approaches, developers create bots for a specific platform, constantly putting developers in the loop. 
Bots do not have the general perceptual capability to understand and navigate with autonomous variability. 
Other architectures in this category, originating from the WoT, attempt to address this challenge by using knowledge models and standards that could enable agents to perceive the web by exposing hypermedia affordances and signifiers~\cite{charpenay6,ciortea2019exploiting}. 
The knowledge descriptions carry discoverable affordances and interpretable signifiers, which can then be resolved by agents~\cite{charpenay6,lemee2022signifiers}. 
This approach might demand extended web standards that make the web a suitable environment for software agents. 
It might also require introducing architectural constraints that web platforms must adhere to in developing and changing their platforms, such as providing a knowledge description where bots can read descriptions of their affordances.

The second category of approaches uses various behavioral cloning and reinforcement learning techniques~\cite{gur2018learning}.
One example is by Shi et al.~\cite{shi2017world}, where they introduce a simulation and live training environment to enable bots to complete web interaction activities utilizing keyboard and mouse actions. 
Recent efforts extend these approaches by leveraging large language models (LLMs) for web page understanding and autonomous web navigation~\cite{gur2022understanding,huang2023language}. 
The results from both techniques and similar approaches reveal the size of the gap between human users and bots~\cite{shi2017world,gur2022understanding}. 

Both approach categories still need to solve the problem of variability and generalizability of perception and action. 
Approaches that leverage the hypermedia knowledge of platforms with affordance and signifier descriptions could serve as placeholders, but real bots with generalizable capabilities would need more autonomous models yet. 

Besides this, some design assumptions consider the environment and the bot as one. 
As a result, they may attempt to design agents as an integrated part of the platforms or try to~\emph{`botify'} and~\emph{`cognify'} or orient web services as agents. 
Alternatively, the whole notion of a~\emph{user-like} bot inherently assumes the bot to have an autonomous presence separate from the web platforms it accesses.
\begin{figure}[!htbp]
\centerline{\includegraphics[scale=0.39]{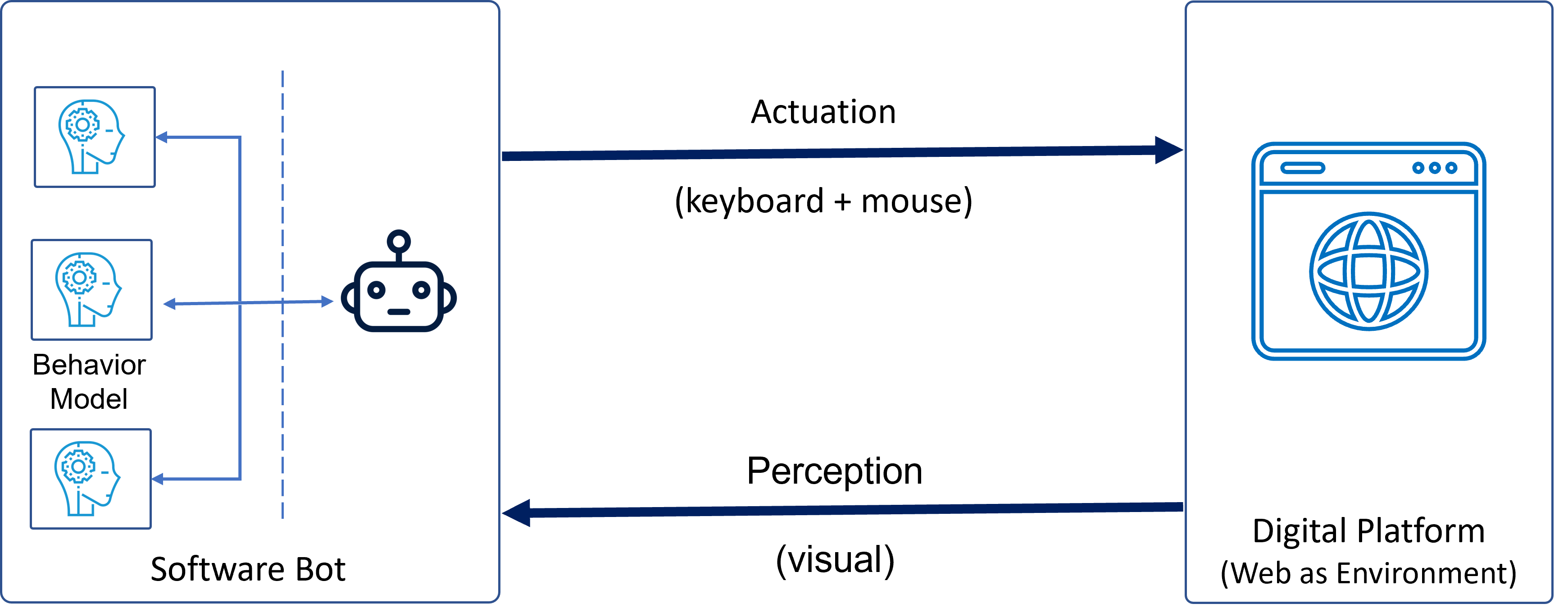}}
\caption{A decoupled bot-environment and bot-behavior~(\emph{left}) viewpoint.}
\label{fig:bot-bm-env}
\end{figure}
Fig.~\ref{fig:bot-bm-env} illustrates the basic perspective in a vertically separate design, the bot, and the web platforms it operates in.  
This strict separation enables both the environment and the bot to evolve independently.
%
%
%
\section{The Behavior Problem}\label{behavior}
Most user activities on digital platforms are complex behaviors resulting from human users' underlying intentions, goals, and belief systems. 
Although a bot operating in digital spaces need not fully emulate humans to achieve generalizable behavior, it is essential to consider the intricacies and sophistication of human users' behavior on the web during bot design~\cite{pennachin2007contemporary}. 
To that end, engineering bots with behavior models similar to human users might take into account existing approaches of measuring generalizable user behavior while not having to replicate human cognition as such~\cite{turing1950computing}.
\begin{figure}[!htbp]
\centerline{\includegraphics[scale=0.48]{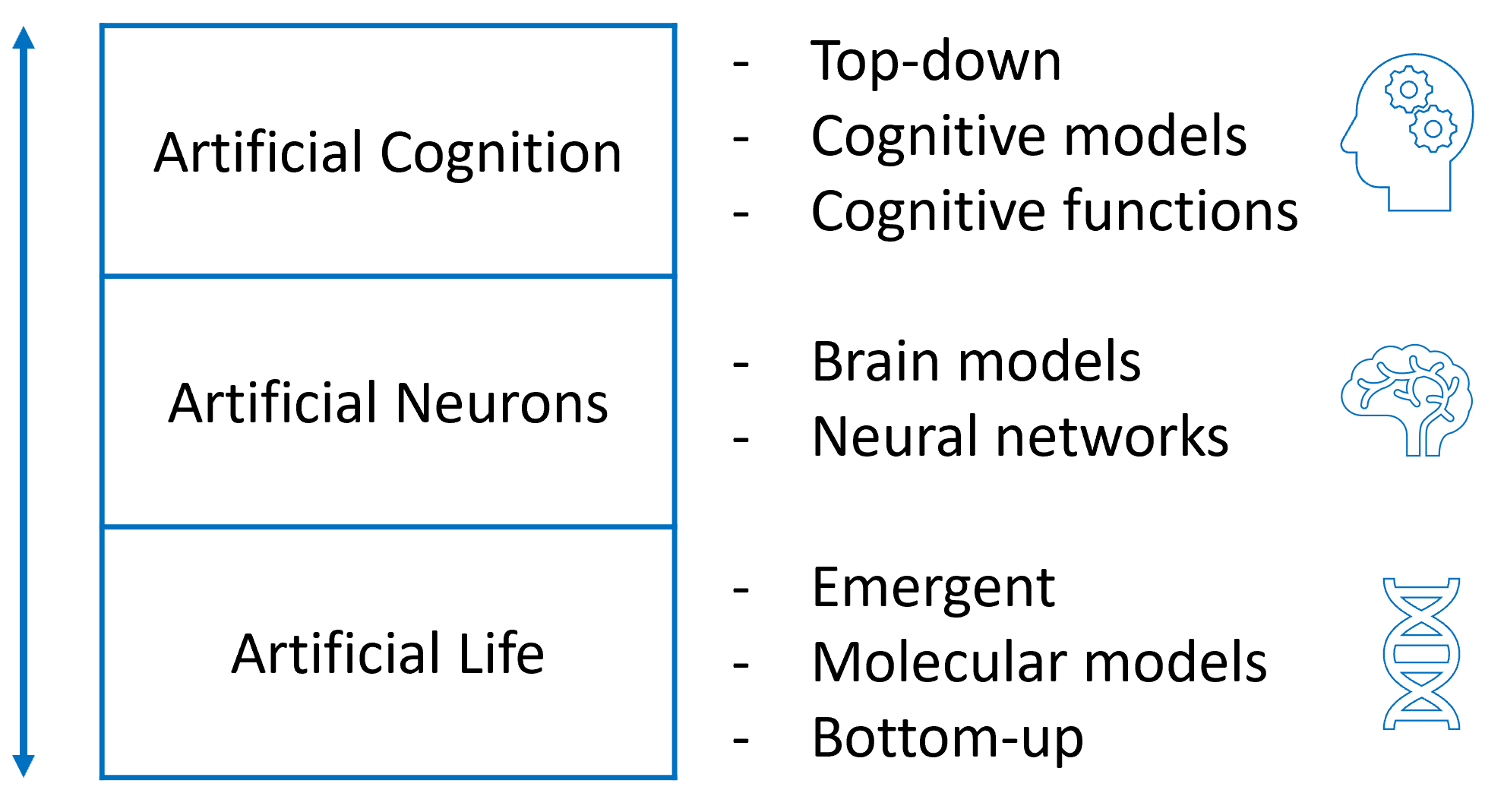}}
\caption{The abstraction ladder in modeling machine intelligence.}
\label{fig:abstractm}
\end{figure}

Current models for engineering intelligent behavior come from three prospective categories of approaches. 
Each approach takes natural or human intelligence as its inspiration and models it at different levels of abstraction. 
The three methods differ mainly in how they try to understand intelligence and where they start the abstraction for modeling intelligence. 
Fig.~\ref{fig:abstractm} illustrates this ladder of abstraction in modeling machine intelligence.
The abstractions start either at artificial cognition, artificial neurons, or artificial life or consciousness~\cite{goertzel2014engineeringPart1,taylor2016webal}. 
These abstractions aim to enact intelligent behavior based, respectively, on high-level cognitive functions, artificial neural networks (ANNs), or more physical and bottom-up approaches starting at molecular or atomic levels.

\textit{Artificial Cognition:}
in cognitive modeling, efforts to model cognition are inspired by the brain's high-level cognitive functions, such as memory. 
Most assumptions are based on studies and understandings in the cognitive sciences. 
Cognitive models use diverse techniques such as production rules, dynamical systems, and quantum models to model particular cognitive capabilities~\cite{schoner2008dynamical}. 
Although cognitive models use methods from other approaches, such as ANNs, they do not necessarily adhere to underlying mechanisms in the brain~\cite{vernonCAs2022,goertzel2014engineeringPart1}. 
Works such as the OpenCog (Hyperon) and the iCub project are promising experimental research examples that heavily rely on artificial cognitive models, i.e., cognitive architectures~\cite{goertzel2014engineeringPart1,vernonCAs2022}.

\textit{Artificial Neurons:} brain models which use artificial neurons aim to understand, model, and simulate underlying computational mechanisms and functions based on assumptions and studies in neuroscience~\cite{fan2019brief}. 
Discoveries from neuroscience are utilized to derive brain-based computational principles. 
Sometimes, these approaches are referred to as~\textit{Brain-derived AI} or~\textit{NeuroAI} models~\cite{walter2021advanced,knoll2019neurorobotics,momennejad2023rubric}. 
Due to the attention given to the underlying principles of computation in the brain, they strictly differ from the brain-inspired cognitive models.
Applications of these models are mainly advancements in artificial neural networks, such as deep learning. 
Large-scale brain simulation research and new hardware development in neuromorphic computing, such as~\textit{SpiNNaker} and~\textit{Loihi}, also contribute to research efforts in this area. 
Some neuromorphic hardware enables close adherence to brain computational principles in particular types of neural networks, such as~\textit{Spiking neural networks}~\cite{maass1997networks,walter2021advanced}. 
Brain-derived AI approaches with neurorobotics aim to achieve embodiment using fully developed morphologies, which are either physical or virtual. 
The Neurorobotics Platform (NRP) is an example of such efforts to develop and simulate embodied systems. 
The NRP is a neurorobotics simulation environment which connects simulated brains to simulated bodies of robots~\cite{knoll2016neurorobotics}.

\textit{Artificial Life (aLife):}
aLife attempts to model consciousness. To do this, researchers and developers start with a bottom-up approach at a physical or molecular level~\cite{taylor2016webal}. 
Most synthesizing efforts to model intelligence in artificial life are simulations with digital avatars.

In the context of bots on web platforms, employing integrated behavior models, such as the NRP and OpenCog mentioned above, is still a challenge. 
Thus, in addition to the proposed separation of the bot and environment, decoupling a bot's basic skeleton and behavior models is architecturally important. 
Fig.~\ref{fig:bot-bm-env},~\textit{left}, illustrates the separate structure of a bot and its behavior models. 
The bot's core skeleton, for example, might have sensory and interaction elements as virtual actuators that enable its operation using the keyboard and mouse actions. 
The vertical separation allows behavior models and bot skeletons to change independently, maintaining the possibility of dynamic coupling. 
%
%
%
\section{Bot Communication Challenges}\label{communication}
In Multi-Agent Systems (MAS), agent-to-agent communication heavily relies on agent communication languages (ACLs) such as FIPA-ACL, standardized by the Foundation for Intelligent Physical Agents(FIPA) consortium~\cite{soon2019review,wooldridge2009introduction,hubner42021,hillmann2014novel}. 
However, in mixed reality environments, where bots and humans share and collaborate in digital spaces, communication cannot rely only on ACLs and APIs~\cite{holz2011mira}.

To that end, a cognitive bot with artificial general intelligence (AGI) must possess communications capabilities to address humans and software agents with diverse communication skills. 
Communication capabilities should include diverse possibilities like email, dialogue systems, voice, blogging, and micro-blogging. 

Large language models (LLMs) have recently shown significant progress in natural language processing and visual perception that could be utilized for bot and human communication~\cite{huang2023language,gur2022understanding}.
%
%
%
\section{Integration and Cooperation Challenges}\label{cooperation}
Researchers assert that the grand challenge in AGI remains in integrating different intelligence components to enable the emergence of advanced generalizable behavior or even collective intelligence~\cite{minsky1988society,goertzel2014engineeringPart1,tononi2015integrated,friston2010free}. 
The intelligence solutions to integrate include learning, memory, perception, actuation, and other cognitive capabilities~\cite{goertzel2014artificial}.
Theories and assumptions developed by proponents include approaches based on cognitive synergy, the free energy principle, and integrated information theory~\cite{goertzel2017toward,friston2010free,tononi2015integrated}. 

In practice, however, integration and cooperation of bots are implemented mainly by utilizing methods such as ontologies, APIs, message routing, communication protocols, and middleware like the blackboard pattern~\cite{dorri2018multi,wooldridge2009introduction,boissier2021autonomous}.

From a software engineering perspective, basic architectural requirements for the context of bots operating on digital platforms are possibilities for the evolvability of bots into collective understanding with shared beliefs, stigmergy, or sharing common behavior models to learn, transfer learned experience, and evolve. 
Other concerns are the hosting, which could be on a cloud or individual nodes, scaling, and distribution of bots and their behavior models. 
\begin{figure}[!htbp]
\centerline{\includegraphics[scale=0.48]{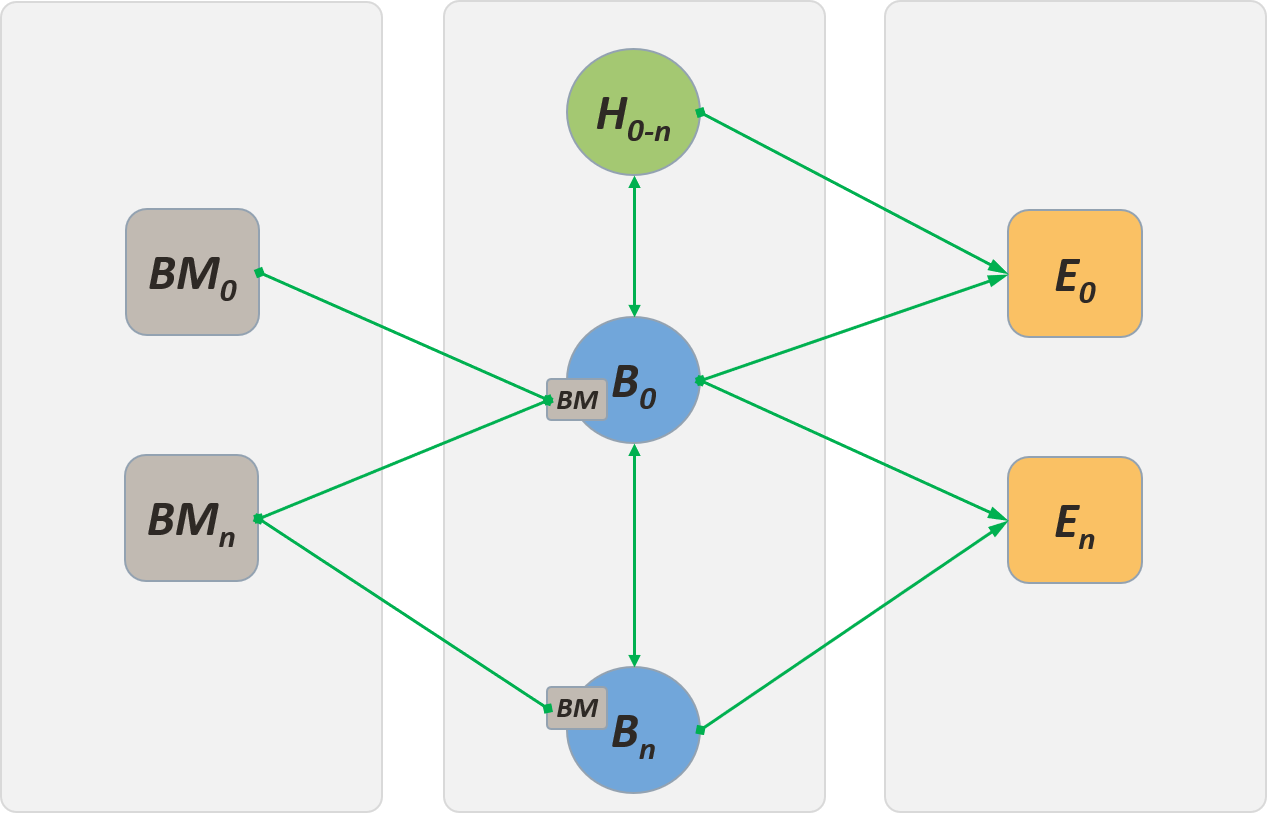}}
\caption{Representation of integrated parts, i.e., bots, shared behavior models, and the web environments.}
\label{fig:botsyss}
\end{figure}

Fig.~\ref{fig:botsyss} shows a simple diagram representing the integrated parts, i.e., bots, shared behavior models, and the environment. 
\emph{B} represents the possible number of bots. 
\emph{BM} represents the shared and individual behavior models.
\emph{E} represents the web environment and its variability. 
The lines represent communication channels.
\emph{H} denotes the human users that participate and share the digital space.
%
%
%
\section{Bot Ethics and Trust}
Concerns and challenges in AGI are diverse.
They touch on various aspects of society and politics and have real-world implications, such as the impact of user-like bots on privacy, security, ethics, and trust with humans~\cite{goertzel2014engineeringEthics,christian2020alignment,coeckelbergh2020aiethics}.
User-like bots, emulating human users' perceptual and interaction techniques, can easily pass bot detection tests and risk exploitation for malicious use cases to deceive and attack web platforms. 
They could also extend their perceptual capabilities beyond the web with connected devices such as microphones and cameras, affecting the personal privacy of others. 
Possible threats include spamming, cyberattacks to overwhelm platforms, and even unfair use of web platforms for surveillance or illicit financial gains.
In WoT context, for instance, bots could affect smart factories and automated services in the real world, compromising physical devices and processes with significant security implications~\cite{kampik2021norms}.

Hypothetically, intelligent social bots could share their beliefs on social platforms similar to or better than any human user, with superb reasoning and argumentation skills.
These cases could negatively impact society by exposing people and software agents to unexpected, misaligned norms and moral beliefs.
Furthermore, deploying advanced cognitive bots as digital workforces may result in unforeseen negative economic consequences. Short-term issues could include unemployment, while long-term concerns may involve ethical dilemmas surrounding bot ownership rights, bot farming, or `enslavement'~\cite{goertzel2014engineeringEthics}.
Accordingly, these ethical concerns may affect the legality of cognitive bot development, potentially impeding their engineering and deployment.
Alternatively, this could introduce new legal aspects regarding regulation, standards, and ethics for integrating and governing bots within emerging socio-technical ecosystems~\cite{kampik2021norms}.

Despite these concerns, bots' current and potential applications can positively impact numerous aspects of society.
Cognitive automation, for example, is driving increased demand for cognitive agents in Industry 4.0, digital twins, and other digital environments~\cite{engel2022cognitive,vogel2020multi,ivanvcic2019robotic}. 
Early implementations, like Wikipedia bots, already play a significant role in fact-checking and other knowledge-based tasks.
On platforms like GitHub, bots assist and automate development tasks~\cite{hukal2019bots}.
Future cognitive bots could also benefit society by participating in knowledge processing and providing valuable new scientific insights, such as medical advancements,  which significantly outweigh their potential risks.

Today, digital platforms handle simple crawling and API-based bots with crawling policies and controlled exposure of APIs. 
However, advanced user-like bots like the ones envisioned in this report will require more complex mechanisms to govern and control their behavior and belief-sharing~\cite{goertzel2014engineeringEthics,kampik2021norms}. 
One approach towards this is ethics and trust by design, which recommends protocols and policies for developers and engineering organizations to incorporate trust models and ethical frameworks at the design and architectural stages~\cite{goertzel2014engineeringEthics}.
Another approach proposes norms and user policies with penalties for agents to acknowledge, understand, and adhere to, similar to what human users would do on digital platforms~\cite{kampik2021norms,kampik2022governance}.
In return, norm and value-aware bots could establish participation, trust, and compliance while facing the consequences of noncompliance. 
They may also contribute to revising and creating collective values and norms, possibly becoming part of  viable socio-technical ecosystems~\cite{kampik2021norms,patrick2002building}.

However, ensuring safety and trust in such ecosystems will require diverse approaches. 
In addition to providing machine-readable norms and policies targeting cognitive agents, it is essential to tackle ethical and trust issues with transparent and explainable design and engineering processes at each stage.
For instance, the European Union (EU) recommends a three-phase human intervention approach at the design phase, at the development and training phase, and at runtime with oversight and override possibilities~\cite{kaur2022trustworthy}. 
As a result, research on developing advanced cognitive bots must also address critical challenges in engineering trustworthy, secure, and verifiable AGI bots employing hybrid approaches. 
%
%
%
\section{Conclusion}
The study presented architectural research challenges in designing and developing a new line of user-like cognitive bots operating autonomously on digital platforms. 
Key challenges, such as the transduction problem, are discussed in the context of digital web platforms' access, user-like visual interaction, and autonomous navigation. 
In the architecture, we recommend bot-environment separation to realize bot autonomy and bot skeleton and behavior model separation for better evolvability. 
Also, bot communication capabilities should include diverse possibilities like email, dialogue systems, and blogging. 
We recommend utilizing shared behavior models for transfer learning or collective intelligence to enact generalizable behavior. 
Finally, we discussed cognitive bots' ethical implications and potential long-term effects, proposing to adopt hybrid approaches that incorporate these aspects into the architecture and the life cycle of bots.

As an outlook, a good starting point for future work would be to conceptualize a detailed implementation architecture and construct a bot by utilizing existing cognitive models. 
These systems can demonstrate the concept and allow further detailed analysis through empirical data and benchmarks. 
%
%
\bibliography{main}
%
\end{sloppypar}
\end{document}